%% file: root.tex
\documentclass[letterpaper, 10 pt, conference]{ieeeconf}  

\IEEEoverridecommandlockouts                              

\title{\LARGE \bf
STAGE: A Stream-Centric Generative World Model for Long-Horizon Driving-Scene Simulation

}

\author{Jiamin Wang$^{1,*}$, Yichen Yao$^{1,*}$, Xiang Feng$^{1}$, Hang Wu$^{2}$, Yaming Wang$^{2}$, Qingqiu Huang$^{2}$\\ Yuexin Ma$^{1,\dag}$, Xinge Zhu$^{2,\dag}$
\thanks{$^{*}$Equal contribution.}%
\thanks{$^{\dag}$Co-corresponding author.}
\thanks{$^{1}$Jiamin Wang, Yichen Yao, Xiang Feng and Yuexin Ma is with ShanghaiTech University, Shanghai, China. Email: {\tt\small \{wangjm2024, yaoych2023, v-fengxiang, mayuexin\}@shanghaitech.edu.cn}.}%
\thanks{$^{2}$Hang Wu, Yaming Wang, Qingqiu Huang, Xinge Zhu is with Yinwang Intelligent Technology Co. Ltd. Email: {\tt\small h.wu@tum.de; wym 24@163.com; huanggingqiu@yinwang.com; \tt\small zhuxinge123@gmail.com}.}%
}

\usepackage{graphicx}
\usepackage{multirow}
\usepackage{amsmath}
\usepackage{amssymb}
\usepackage{hyperref}

\begin{document}

\maketitle
\thispagestyle{empty}
\pagestyle{empty}

\begin{abstract}

  \input{Sections/0.Abstract}

\end{abstract}

\section{INTRODUCTION}
    \input{Sections/1.Introduction}

\section{RELATED WORKS}
    \input{Sections/2.RelatedWorks}

\section{METHODS}
    \input{Sections/3.Method}
\vspace{-30pt}
\section{EXPERIMENT}
    \input{Sections/4.Experiment}

\section{CONCLUSION}
    \input{Sections/5.Conclusion}










\bibliographystyle{ieeetr}
\bibliography{iros2025}
\newpage

\end{document}

%% file: Sections/0.Abstract.tex
The generation of temporally consistent, high-fidelity driving videos over extended horizons presents a fundamental challenge in autonomous driving world modeling. Existing approaches often suffer from error accumulation and feature misalignment due to inadequate decoupling of spatio-temporal dynamics and limited cross-frame feature propagation mechanisms. To address these limitations, we present STAGE (Streaming Temporal Attention Generative Engine), a novel auto-regressive framework that pioneers hierarchical feature coordination and multi-phase optimization for sustainable video synthesis.
To achieve high-quality long-horizon driving video generation, we introduce Hierarchical Temporal Feature Transfer (HTFT) and a novel multi-stage training strategy. HTFT enhances temporal consistency between video frames throughout the video generation process by modeling the temporal and denoising process separately and transferring denoising features between frames. The multi-stage training strategy is to divide the training into three stages, through model decoupling and auto-regressive inference process simulation, thereby accelerating model convergence and reducing error accumulation.
Experiments on the Nuscenes dataset show that STAGE has significantly surpassed existing methods in the long-horizon driving video generation task. In addition, we also explored STAGE's ability to generate unlimited-length driving videos. We generated 600 frames of high-quality driving videos on the Nuscenes dataset, which far exceeds the maximum length achievable by existing methods. Our project homepage:\url{https://4dvlab.github.io/STAGE/}

%% file: Sections/1.Introduction.tex
World model summarizes historical information to capture a comprehensive representation of the environment and forecasts future states~\cite{path,worldsurvey}. 
This capability of the world model is particularly important for video generation because video generation requires summarizing physical laws from historical frames to predict the next video frame. In autonomous driving, many tasks are data-intensive, requiring large amounts of annotated data to train models. Moreover, annotating large-scale, real-world driving scenarios is expensive. Therefore, world models for generating driving videos are of critical importance, especially for generating high-quality long-horizon video sequences. High-quality images provide richer details, allowing autonomous driving agents to better recognize and interpret their environment. Meanwhile, generating long-horizon videos provides a controlled and stable environment for extended training and testing, reliance on costly, real-world data collection.

\begin{figure}[t]
  \includegraphics[width=\linewidth]{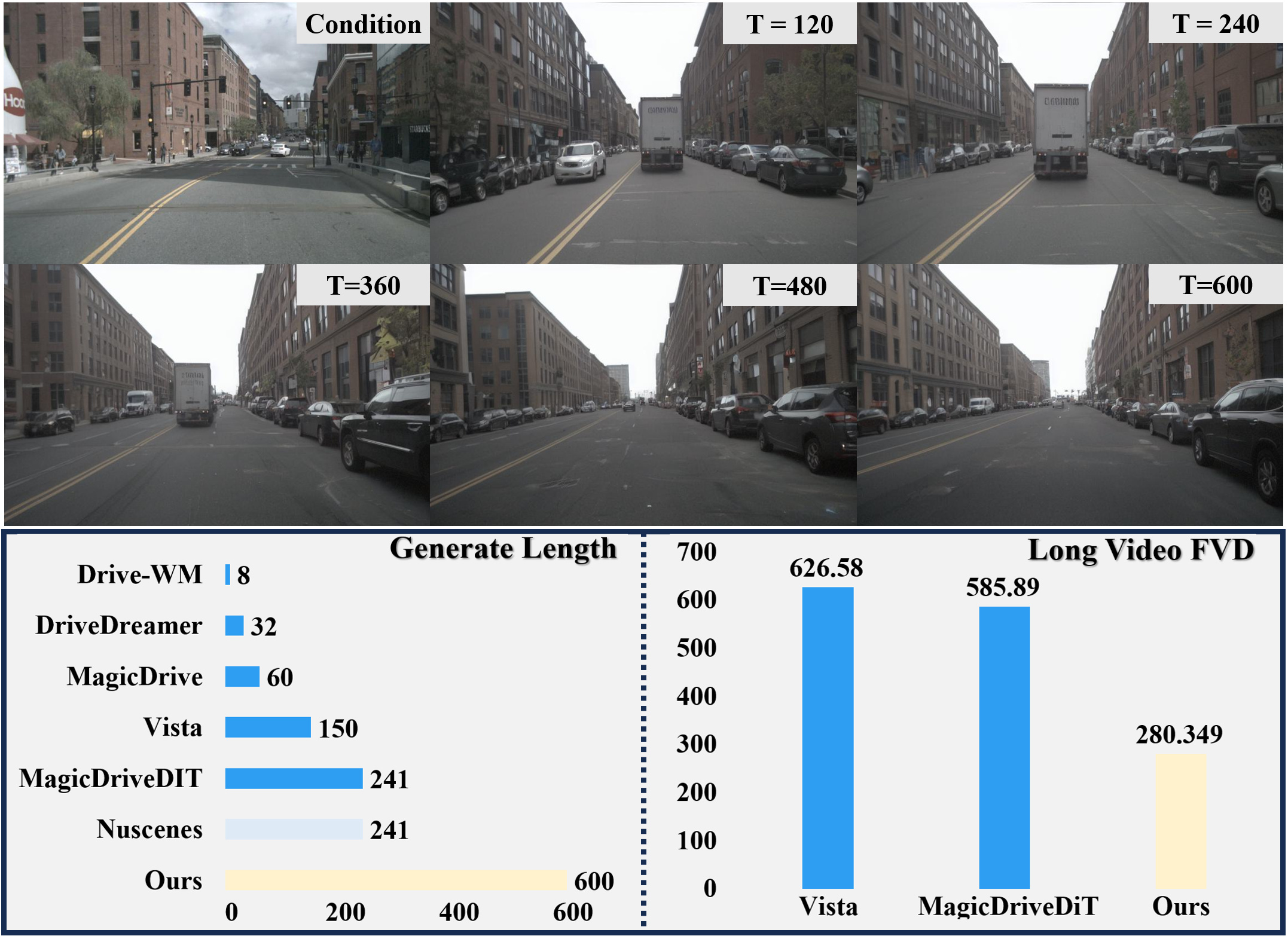}
  \vspace{-18pt}
  \caption{STAGE creates extended videos that exceed the original length of Nuscenes data, outperforming other methods in both duration and performance. }
  \vspace{-15pt}
\label{fig:pipeline}
\end{figure}

However, traffic scenes involve a diverse range of participants, making them highly complex and dynamic. Each participant operates within distinct physical constraints, which further complicates the task of modeling these interactions. As a result, capturing the complex and dynamic physical laws in such environments becomes challenging, making it difficult to generate participants with realistic movements and geometric shapes. Furthermore, generating high-quality long-horizon video sequences imposes additional demands on computational resources, driving scene details, and the temporal consistency between driving video frames. To generate high-quality long-horizon driving videos, some works~\cite{magicdrivedit,magicdrive} leverage diffusion models for one-shot generation, such as MagicDriveDiT~\cite{magicdrivedit}. However, these methods cannot flexibly adjust the number of generated frames, nor can they generate an infinite number of frames. In addition, there are the following methods~\cite{vista,drive-WM,xie2025glad} that adopt auto-regressive architectures, which generate long videos through multiple inference steps. Each inference step generates a short video chunk conditioned on the last few frames of the previous chunk. As a special form of auto-regressive architecture, the streaming auto-regressive model ~\cite{streamingt2v} is generated frame by frame and can more flexibly generate videos of any length. Similar to the aforementioned methods, the above auto-regressive generation method will accumulate errors during the auto-regressive process, which leads to serious degradation of video quality during the video length expansion process.


To address these limitations 
mentioned above
, we proposed STAGE, a Streaming Temporal Attention Generative Engine For Driving-Scene Simulation. It is a novel streaming-based generative world model for autonomous driving, which flexibly generates high-quality videos of arbitrary length through frame-by-frame generation. To generate high-quality long-horizon driving videos, we introduce Hierarchical Temporal Feature Transfer (HTFT) and a novel multi-stage training strategy. HTFT models both the temporal and denoising processes separately, transmitting the denoising features of previous frames to the current frame during generation. Information transfer between frames at the same denoising step enhances temporal consistency across video frames throughout the video generation process. The multi-stage training strategy divides the training into three stages: the first two decouple HTFT from the model to accelerate convergence, while the third addresses the issue of training-inference discrepancy by simulating scenarios where the model encounters its own generated data during auto-regressive inference, thereby reducing error accumulation.


In addition, we explore the possibility of STAGE for generating longer video sequences, as shown in Figure \ref{fig:pipeline}. By training solely on the NuScenes~\cite{qian2024nuscenes} dataset, we successfully extend the video length, enabling the generation of stable, high-quality video sequences up to 600 frames, thus providing a foundation for theoretically unlimited video generation.

In summary, our contributions are as follows.
   \begin{itemize}
       \item We present STAGE, a novel streaming-based generative world model for autonomous driving, which flexibly generates high-quality videos of arbitrary length through frame-by-frame generation.
       \item We proposed a strategy for feature transfer named HTFT, which enhances temporal consistency between video frames throughout the video generation process by modeling the temporal and denoising process and transferring denoising features between frames.
       
       \item We propose a novel multi-stage training strategy, through model decoupling and auto-regressive inference process simulation, thereby accelerating convergence and reducing error accumulation.
       
       \item In the task of long-horizon video generation for autonomous driving, STAGE achieves state-of-the-art performance, significantly outperforming other approaches in both length and quality.
   \end{itemize}

%% file: Sections/2.RelatedWorks.tex
\subsection{World model for Autonomous Driving}
World models predict future states based on historical observations, offering a promising approach for data generation in autonomous driving. Recent models~\cite{bevcontrol, BEVGen, gaia, genad} have focused on driving video generation. GAIA-1~\cite{gaia} and GenAD~\cite{genad} use diffusion models to generate future scenes but struggle with stability due to lack of layout control. Later models~\cite{magicdrive, wang2023drivedreamer, drivingdiffusion, panacea} generate short videos (8-25 frames) using control signals and encoding strategies.

The above-proposed models still struggle to overcome quality and generation length limitations. Generating high-quality long-horizon driving videos remains a challenge. Recently, auto-regressive and one-shot generating methods have been adopted to achieve that. MagicDriveDiT~\cite{magicdrivedit} uses a one-shot method to generate a 240-frame video via the DiT architecture~\cite{DIT}, which is resource-intensive, particularly in terms of VRAM, and lacks flexibility in adjusting video length, making closed-loop simulation costly. For auto-regressive methods~\cite{vista, drive-WM,xie2025glad,dreamforge}, they can generate long videos through multiple inferences. However, due to the accumulation of errors in the auto-regressive process, the video quality deteriorates. In contrast, our method generates high-quality long videos in a frame-by-frame manner, which offers greater flexibility. Meanwhile, in theory, it allows for the generation of videos of unlimited length.




\subsection{Long Video Generation}
Long video generation is a challenge due to temporal complexity and resource constraints. Previous work~\cite{latent, NUWA-XL,longvideogen} has done a lot of research in this field.
However, they encounter challenges in flexibly adjusting the generation length and generating infinitely.

Recently, to overcome the temporal constraints, FIFO-diffusion~\cite{fifo} employs a training-free diagonal denoising approach to propagate the current context across subsequent frames, enabling the generation of consistent videos.
StreamingT2V~\cite{streamingt2v} utilizes auto-regressive approach with the injection of anchor frames and long-short memory modules to maintain temporal coherence. 
However, these methods 
are not designed to generate videos with explicit controllability. Compared with them, our method focuses on generating complex and realistic driving scenes generation with precise layout control, such as Bounding Box, HDmap, etc. 


\begin{figure*}[ht]
  \includegraphics[width=\linewidth]{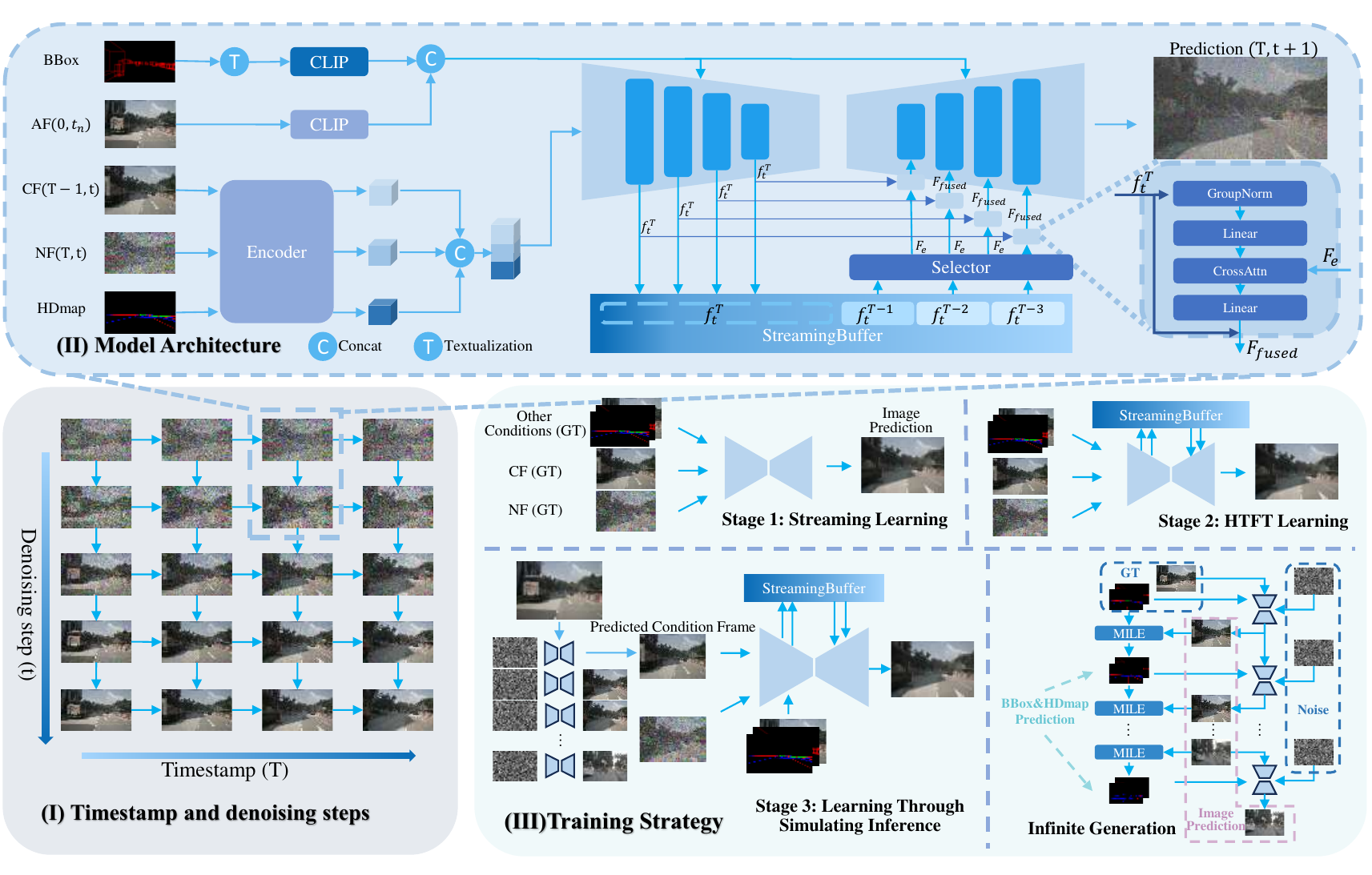}
  \vspace{-22pt}
  \caption{\textbf{Pipeline of STAGE.} “AF”, “CF”, “NF” stand for Anchor Frame, Condition Frame, Noise Frame, respectively. (\uppercase\expandafter{\romannumeral1}) illustrates the hierarchical structuring of time and denoising steps, with the horizontal axis representing time and the vertical axis representing the denoising steps. $T$ represents the $T$-th frame, while $t$ refers to the $t$-th denoising step. 
  (\uppercase\expandafter{\romannumeral2}) illustrates the framework of our model, where we leverage HTFT to facilitate feature transfer along the temporal dimension, thereby refining the generation process at each step. (\uppercase\expandafter{\romannumeral3}) presents our multi-stage training strategy and the process for infinite generation.}
  \vspace{-10pt}
\label{fig:pipeline}
\end{figure*}

%% file: Sections/3.Method.tex
In this section, we present our STAGE method for driving video generation. In Section~\ref{problem}, we outline the description of the problem and the overall approach of the STAGE method. Section~\ref{Hierarchical} focuses on Hierarchical Temporal Feature Transfer, which improves temporal consistency in the video generation process by facilitating feature transfer between frames. In Section~\ref{Training_strategy}, we propose a multi-stage training strategy that addresses the inconsistency between training and inference by simulating unseen self-generated data during auto-regressive inference. Lastly, in Section~\ref{Infinite_Generation}, we present an infinite generation method, where predicted condition information is used to guide the generation of subsequent frames.

\subsection{Problem Formulation and Overview}
\label{problem}

This paper aims to generate high-quality, controllable, long-horizon driving videos. Given the initial frame $S_0$, our goal is to generate a coherent, temporally consistent video sequence $\{I_{0}, I_{1}, ..., I_{T}\}$, where $I_{t}$ is the $t$-th frame and $T$ is the video length. We use the HDMap $H_0$, Bounding Box $B_0$, and initial frame image $I_0$ as input.  Figure \ref{fig:pipeline} illustrates the overall framework of our model. The entire framework generates latent features for future frames in a streaming manner, progressively producing the latent feature for each subsequent frame. Specifically, at the $T$-th frame and $t$-th denoising step, we input the latent feature $z_{t}^{T}$, predicted HD map, bounding box description, condition frame (latent feature of $I_{T-1}$), and anchor frame (latent feature of $I_0$). We use the StreamingBuffer to store features and apply Hierarchical Temporal Feature Transfer (HTFT) to ensure temporal consistency across frames.

\subsection{Hierarchical Temporal Feature Transfer}  
\label{Hierarchical}

Hierarchical Temporal Feature Transfer (HTFT) is designed to preserve temporal consistency across video frames during the generation process. Previous methods typically rely on temporal attention modules or pre-trained video generation models to ensure frame consistency. However, these approaches are not directly applicable to our method, which generates frames sequentially. Stable Diffusion~\cite{rombach2022high}, being a powerful pre-trained image-generation model trained on large-scale datasets, aligns well with our task and serves as the foundational framework for our approach. Maintaining temporal consistency throughout the generation process is a key challenge. Recognizing that the features within the U-Net in Stable Diffusion provide additional contextual information, which aids in recovering real images from noise during the decoding phase, we introduce HTFT to enhance temporal feature transfer. During the denoising process, we integrate not only the features of the current frame but also those from preceding frames, offering guidance in both spatial and temporal dimensions during the decoding phase.

Specifically, for the denoising process $x_t^T \rightarrow x_{t+1}^T$ at denoising step $t$ of frame $T$, we select the features of the previous \( K \) frames from the StreamingBuffer to obtain the fused feature. The process is described as follows:
\begin{align*}
F_{s} &= \text{FIFO}\left(\{ f_{t}^{T-1}, f_{t}^{T-2}, \dots, f_{t}^{T-N} \}\right), \\
F_{e} &= \bigcup_{x \in S} F_{s}[x], \\
g_{t}^{T} &= Linear(GroupNorm(f_{t}^{T})), \\
H_{t}^{T} &= CrossAttn(g_{t}^{T},F_{e}), \\
F_{fused} &= Dropout(f_{t}^{T}) + Linear(H_{t}^{T}),
\end{align*}
where $F_{s}$ is a FIFO-queue with fixed length of $N$ which stores the features of the previous frames. $S$ is a predefined set containing the indices of the features that need to be selected from the StreamingBuffer. $F_{fused}$ represents the final fused feature, which serves as the guiding signal for the generation process.


 
In our method, we set $N$ = 10 and $S$ = \{-1, -5, -10\}, meaning that features from the $1^{st}$, $5^{th}$ and $10^{th}$ previous frames are extracted from the StreamingBuffer at each step. During training, a streaming data sampler samples frames in chronological order to preserve temporal coherence. The same noise intensity is applied to frames within a scene to ensure consistent denoising. Additionally, we employ a StreamingBuffer mechanism that stores features from previous frames. This approach avoids redundant feature computation, thereby improving training efficiency. 
  
To enhance training efficiency, we introduce the StreamingBuffer, a fixed-length FIFO cache queue that stores features extracted from previous frames. 
When predicting the latent representation of the next frame, more temporally distant features can integrate additional information but also bring higher computational costs. To balance performance and computational expenses, while considering the similarity of information between frames, we employ a frame-skipping strategy. It selects $K$ frames from the previous $N$ frames in the StreamingBuffer for feature fusion with the U-Net model.
The StreamingBuffer is updated with the most recent frame features. During the inference phase, a two-dimensional StreamingBuffer is maintained for each denoising step. When transitioning between these steps, the relevant one-dimensional StreamingBuffer is utilized and updated, ensuring the continuous integration of prior frame information at each step.

\subsection{Training Strategy}
\label{Training_strategy}
We propose a three-stage training scheme that progressively equips the model with robust long-horizon sequence generation capabilities, effectively addressing the training-inference mismatch inherent in streaming approaches. Our training strategy are illustrated in Fig.~\ref{fig:pipeline}. Furthermore, we incorporate data augmentation techniques to enhance the model’s generalization ability.

\subsubsection{Stage 1: Streaming Learning}
In the first stage, we focus on developing a video-generation-based world model equipped with streaming and conditional-follow capabilities. By initializing the model with pre-trained weights and training it across all conditional inputs, we ensure foundational functionality while disabling the HTFT (Hierarchical Temporal Feature Transformer) and its associated network components during this phase.

\subsubsection{Stage 2: HTFT Learning}
After the first stage, we obtain a model with basic streaming ability. During this stage, we utilize HTFT to provide refined and temporally informed guidance during generating enhance temporal consistency across video frames during the generation process. 

The transfer in HTFT has two dimensions: time and the number of denoising iterations. To obtain the feature used for transfer at time $T$ and denoising iteration $t$, at least $T$ inferences are required because every feature at each time step depends on features from the previous time steps with the same denoising iteration. To reduce the training cost, we froze all the parameters before HTFT and train the model sequence by sequence. Within each epoch, we fix the denoising iteration for each sequence, enabling us to cache features for the forward pass without performing costly repeated inferences.

\subsubsection{Stage 3: Learning Through Simulating Inference}
To mitigate the mismatch between training and inference and to enhance the model’s generative ability, we replace the condition frame with its inferred version. Similar to Stage 2, we train the model on a sequence-by-sequence basis. For each sequence, we first generate an image for each frame through inference. We then replace the condition frame with the inferred image to simulate image quality degradation during inference. Note that we perform inference once per sequence, rather than for each frame, to reduce computational cost to a reasonable level.

\subsubsection{Data Augmentation Strategy}
To mitigate the degradation of image quality during the generation process and enhance the stability of the generated outputs, we apply a Discrete Cosine Transform filtering technique to the condition frame. This method randomly eliminates a portion of the high-frequency components, simulating the discrepancies between the generated images and the ground truth. Additionally, we incorporate random dropout and introduce small amounts of noise to the conditioning input, thereby improving the model's robustness and its ability to generalize across variations in the data.

\subsubsection{Loss}
We use the L2 distance as our loss function for noise prediction.

\begin{equation}
L = \mathbb{E}_{x_{t},x_0,t } \left[ \| W_{aux} \odot (x_{0} - \epsilon_\theta(x_t, t)) \|^2 \right].
  \label{eq:mse_loss} 
\end{equation}
To highlight foreground objects (e.g., vehicles and pedestrians), we assign auxiliary weights $W_{aux}$ at the loss level. Furthermore, by computing the size of each foreground object, we provide higher weights to smaller ones to improve the model’s ability to generate distant and small objects. Unlike bounding boxes used in~\cite{chengeodiffusion}, we use convex hulls for a more precise geometric representation and area calculation. The Convex hull is calculated through 8 corner points of the 3D bounding box. Because size is used to balance weights for smaller objects, we only consider the polygon area within the camera view when calculating object sizes.

\begin{equation}
\begin{gathered}w_{i j}^{\prime}=\left\{\begin{array}{lc}k / p_{i j}^c & (i, j) \in \text { foreground polygon } \\ 1 /\left(H * W\right)^c & (i, j) \in \text { background polygon },\end{array}\right. \\ w_{i j}=H * W * w_{i j}^{\prime} / \sum w_{i j}^{\prime},\end{gathered}
\label{eq:mse_loss} 
\end{equation}
where $c$ and $k$ are tunable constants, $p_{i j}$ denotes the size of the convex hull containing the pixel and the image latent has dimensions $H$,$W$.

\subsection{Infinite Generation}
\label{Infinite_Generation}
When the required video length exceeds that of the NuScenes\cite{qian2024nuscenes} scene video, the necessary conditions such as HD map and Bounding Box annotations are unavailable. Therefore, we need a model capable of predicting future conditions. Inspired by MILE~\cite{hu2022model}, we integrate its framework into our model, as illustrated in Fig.~\ref{fig:pipeline}. and trained it on the NuScenes dataset. Specifically, at frame $T$, we use MILE to predict the ego vehicle's orientation $O_{t+1}$, velocity $V_{t+1}$, acceleration $A_{t+1}$, and the labels $L_{t+1}$ from the forward BEV view. We then extract lane boundaries and pedestrian crossings from $L_{t+1}$ to form the HD map, and human and vehicle labels as Bounding Boxes. Using the orientation $O_{t+1}$, the coordinate transformation matrix $T_{ego2cam}$ from the ego vehicle to the front-view camera, and the camera intrinsic matrix $K$, we project the HD map and Bounding Boxes onto the front-view. These newly generated conditions will guide STAGE in predicting the image corresponding to the next frame $I_{T+1}$.

\begin{figure*}
  \includegraphics[width=\linewidth]{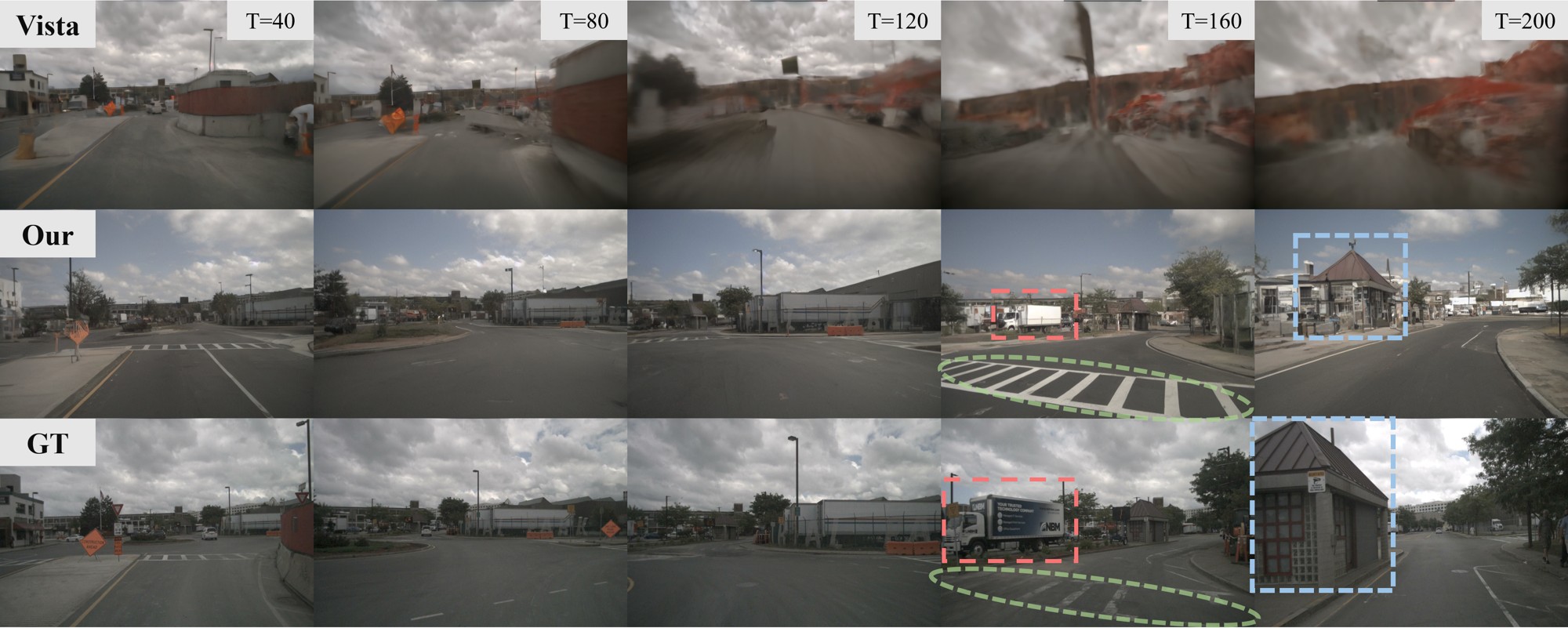}
  \vspace{-18pt}
  \caption{\textbf{Qualitative Comparison between Vista and STAGE in long video generation task.} We generated 201 frames and selected frames 41, 81, 121, 161, and 201 for the comparison.}
    \vspace{-5pt}
\label{fig:long video}
\end{figure*}

\begin{figure*}[ht]
  \includegraphics[width=\linewidth]{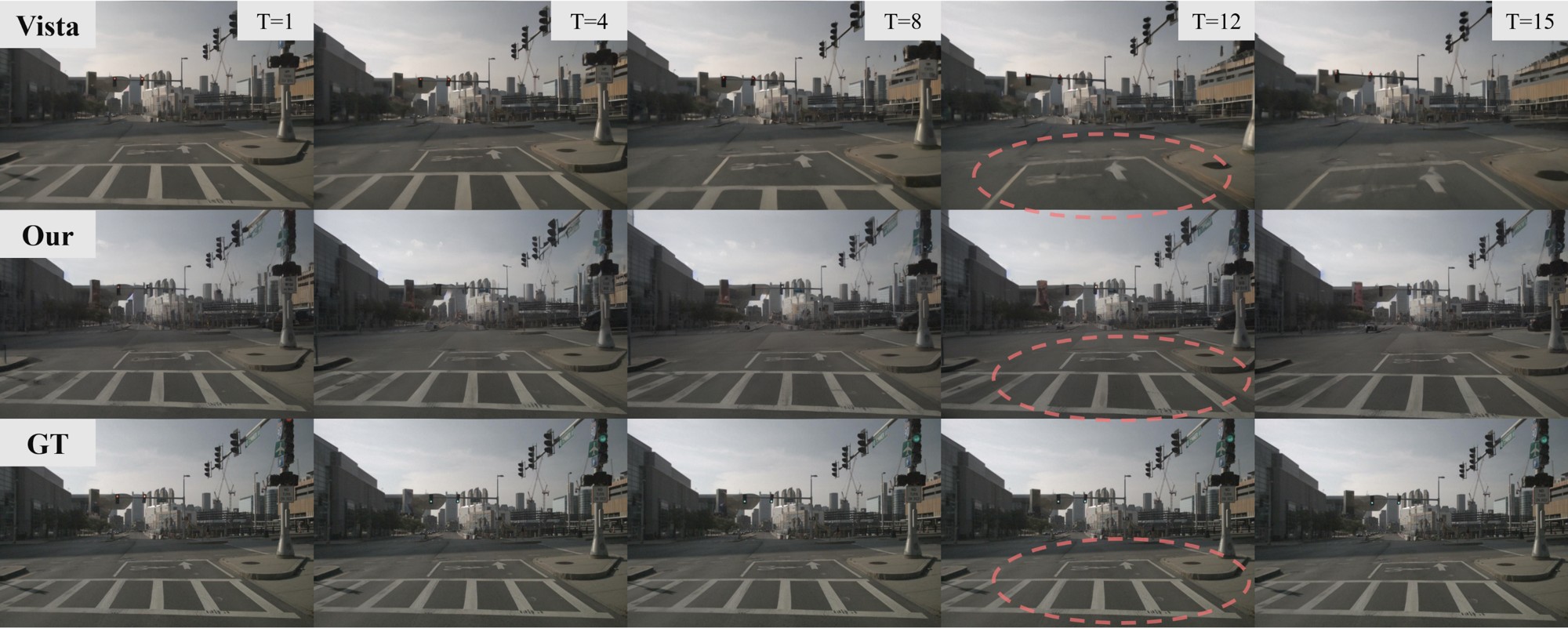}
  \vspace{-18pt}
  \caption{\textbf{Qualitative Comparison between Vista and STAGE in short video generation task.} We generated 16 frames, and selected frames 2, 5, 9, and 16 for comparison. }
\label{fig:short_video}
\end{figure*}

\begin{figure*}
  \includegraphics[width=\linewidth]{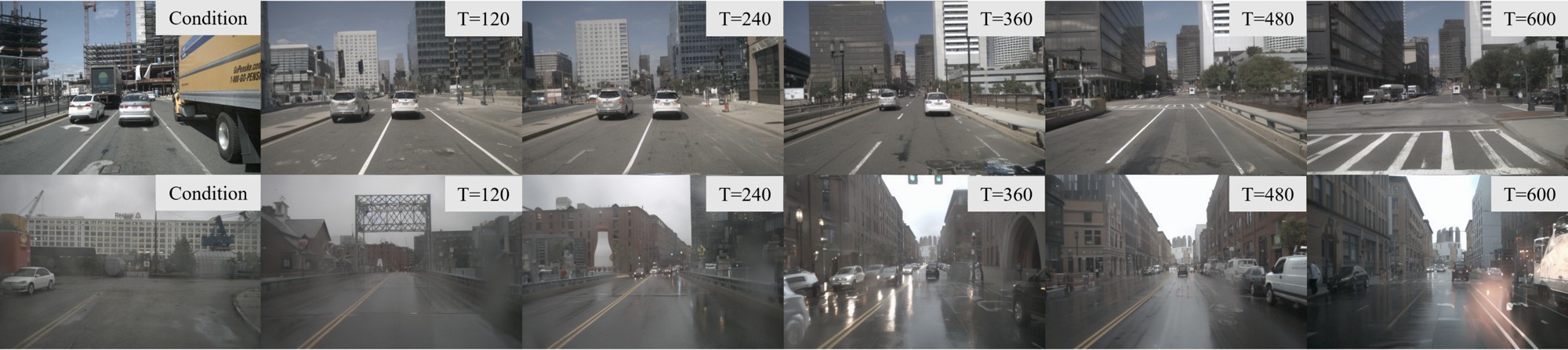}
  \vspace{-18pt}
  \caption{\textbf{Visualization of Longer Video Generation.} We generate 601 frames and selected frames 121, 241, 361, 481, and 601 for the visualization.}
  \vspace{-10pt}
\label{fig:longer video}
\end{figure*}

\begin{figure}
  \includegraphics[width=\linewidth]{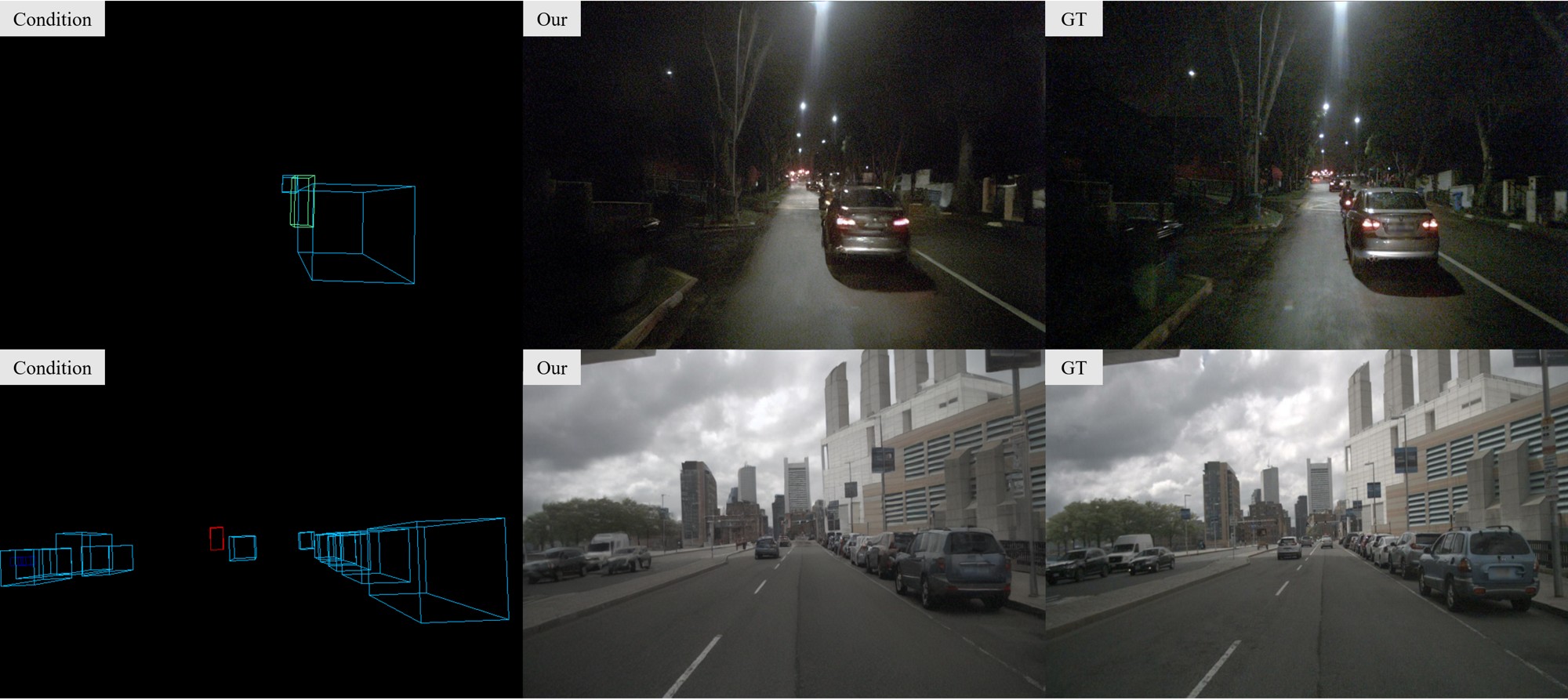}
  \vspace{-18pt}
  \caption{\textbf{Visualization of Bounding Box Control.} On the left is the bounding box provided by the ground truth, in the middle are the results generated by our model, and on the right are the ground truth images.}
  \vspace{-10pt}
\label{fig:bbox}
\end{figure}
\vspace{1.2cm}


%% file: Sections/4.Experiment.tex
\subsection{Datasets and Implementation Detail}

\textbf{Datasets}. The training data comes from the real-world driving dataset NuScenes~\cite{qian2024nuscenes}, which consists of 700 training and 150 validation sequences, each around 20 seconds in duration. Following the approach of DriveDreamer~\cite{wang2023drivedreamer}, we have extended the annotation frequency to 12 Hz. 

\textbf{Implement Details}. Our STAGE model is initialized with pre-trained weights from Stable Diffusion v1.4~\cite{rombach2022high}. We replicate cross-attention weights at each block layer to guide image generation using the anchor frame's latent features. All experiments were conducted on 80 Ascend 910B devices for training. We spent 2 days on Stage 1, 1 day on Stage 2, 4 days on Stage 3, and 2.5 days on Stage 4 for the training process. In both Stage 1 and Stage 2, the learning rate was set to 5e-5, while in Stage 3, it was reduced to 1e-5. During inference, we employed the EDM~\cite{karras2022elucidating} sampler to perform 64 sampling steps, generating images at a resolution of $768 \times 512$.

\subsection{Experiment Details}
\textbf{Evaluation Metrics}. 
We use FVD~\cite{FVD} and FID~\cite{FID} to evaluate video and image quality, respectively, with VideoGPT~\cite{Videogpt} for FVD extraction and Inception-v3~\cite{inception-v3-model} for FID computation.


\textbf{short-horizon Generation Setting} We evaluate our model on the entire nuScenes validation set, providing a ground-truth condition frame every 16 frames. For FVD, each segment has 16 frames. To ensure more accurate results, we generate four sets of outputs, combine them, and then compute both FID and FVD. The FID and FVD values of other methods are taken from their respective original papers.
For FVD, we use VideoGPT as our feature extractor.

\textbf{long-horizon Generation Setting}
Similar to~\cite{magicdrivedit}, we randomly sample 60 sequences from the nuScenes validation set and report FVD and FID. Each video is divided into 16-frame segments for FVD computation. Results for MagicDriveDiT~\cite{magicdrivedit} are taken from the original papers, while results for Vista~\cite{vista} are obtained from the official checkpoint.

\begin{table}
\centering
\caption{Comparison with state-of-the-art methods in short-horizon and long-horizon generation. Lower is better for all metrics.}

\label{lab:short-long-horizon}
\begin{tabular}{llll}
\hline
\multicolumn{1}{l}{Setting} & \multicolumn{1}{l}{Methods} & \multicolumn{1}{l}{FID↓} & \multicolumn{1}{l}{FVD↓} \\ \hline
\multirow{7}{*}{short term} & DriveDreamer \cite{wang2023drivedreamer}               & 52.60                    & 452.00                   \\
                            & Drive-WM \cite{drive-WM}                   & 15.80                    & 122.70                   \\
                            & MagicDrive \cite{magicdrive}                 & 16.20                    & 217.90                   \\
                            & MagicDriveDiT \cite{magicdrivedit}              & 20.91                    & \textbf{94.84}                    \\
                            & DreamForge \cite{dreamforge}                & 16.00                    & 224.80                   \\
                            & Ours                        & \textbf{11.04}                    & 242.79                   \\ \hline
\multirow{3}{*}{long term}  & MagicDriveDiT  \cite{magicdrivedit}             & -                        & 585.89                   \\
                            & Vista  \cite{vista}                      & 90.55                    & 626.58                   \\
                            & Ours                        & \textbf{23.70}                    & \textbf{280.34}                   \\ \hline
\end{tabular}

\end{table}

\begin{table}[!t]
\centering
\caption{The ablation studies at different training stages. Lower is better for all metrics.}

\label{tab:abl}
\begin{tabular}{lll}
\hline
     & FID↓     & FVD↓      \\ \hline
Stage 1 & 17.09  & 508.29 \\
Stage 2 & 11.90 & 245.11 \\
Stage 3 & \textbf{11.04} & \textbf{242.79} \\ \hline
\end{tabular}
\end{table}
\subsection{Results and analysis}

\textbf{Quantitative results.} We present a comparison with other methods on the NuScenes dataset in Table.\ref{lab:short-long-horizon}, focusing on two main aspects: long-horizon sequence generation and short-horizon sequence generation. In the long-horizon sequence generation task, our model achieves state-of-the-art performance with lower FID and FVD. Compared to current state-of-the art (SOTA) method MagicDriveDiT, our method reduces the FVD by half. This improvement is due to the stability gained from the streaming inference, which generates frames progressively. Additionally, it is attributed to our multi-stage training strategy, which simulates the degradation of image quality during inference and mitigates its decline during training.

In the short-horizon sequence generation task, our model achieves the lowest FID; however, its FVD is relatively higher compared to other methods. This can be attributed to the fact that most previous methods are based on the Stable Video Diffusion~\cite{blattmann2023stable} is desigend for short video generation and is pre-trained on large-scale short video datasets. 

Furthermore, by comparing the FVD scores for both short-horizon and long-horizon generation tasks, our method exhibits only minimal performance degradation, highlighting the long-horizon temporal stability of our model. This further validates the effectiveness of our approach.


\textbf{Qualitative results.} Fig.~\ref{fig:long video} shows the comparison between Vista~\cite{vista} and STAGE in long video generation. After multiple auto-regressive inferences, the image quality of Vista gradually deteriorates due to the accumulation of errors. In contrast, our STAGE demonstrates significant stability during the generation process. Due to the use of only anchor frame, bounding boxes, and HD map annotations as conditions during inference, there are some differences from the ground truth in long-horizon video generation tasks, as seen in the house, vehicles, and pedestrian crossing circled in Fig.~\ref{fig:long video}. This also demonstrates the diversity of images generated by STAGE.

Fig.~\ref{fig:short_video} shows a comparison between Vista and STAGE in short video generation. Both models produce images of high quality, but the ego car in Vista's video moves forward, whereas it remains stationary in both the ground truth and our results. This indicates that STAGE has stronger command-following ability.

\subsection{Ablation Study}

To validate the effectiveness of the proposed training strategy, we conducted ablation experiments on short-horizon sequence generation tasks. Table.~\ref{tab:abl} presents the results of the ablation study. It can be observed that Stage 2, which incorporates the HTFT strategy, shows significant improvements in image quality and consistency of the video sequence. HTFT plays a crucial role in enhancing both. In Stage 3, by maintaining consistency between training and inference, the FID and FVD performance metrics exhibit a noticeable decline.

\subsection{Generation Exceeding Original Data Length}
Our method generates videos frame by frame, theoretically allowing for generating infinite frames. To showcase the model's capability for generating super long sequences, 
we employ the stage 4 model to generate a continuous sequence of 600 frames, as illustrated in Fig.\ref{fig:longer video}. Notably, this greatly surpasses the length of sequences present in both the training and validation datasets of NuScenes~\cite{qian2024nuscenes}. Meanwhile, We observe negligible degradation in image quality over time., highlighting the method's potential for infinite frame generation. Note that, apart from the initial frame, all conditioning frames used for generation in this demonstration are predictions obtained directly from the stage 4 model.

\subsection{Condition Follow}
We utilize GT bounding boxes as input to showcase our model's condition-following ability. As shown in Fig.\ref{fig:bbox}, our model effectively generates vehicles at the specified locations, demonstrating impressive condition-following capability both during the day and at night.

%% file: Sections/5.Conclusion.tex

In this paper, we introduce STAGE, a novel streaming-based generative world model for autonomous driving. STAGE generates high-quality videos of arbitrary length in a frame-by-frame manner, enabled by innovative temporal feature transfer and a progressive training strategy. Leveraging this novel design, STAGE is capable of producing high-quality long-horizon driving videos. Our method achieves state-of-the-art (SOTA) performance on public datasets, significantly outperforming existing methods. Furthermore, our experiments demonstrate the possibility of generating high-quality autonomous driving videos of infinite length.

%% file: root.bbl
\begin{thebibliography}{10}

\bibitem{path}
Y.~LeCun, ``A path towards autonomous machine intelligence version 0.9. 2, 2022-06-27,'' {\em Open Review}, vol.~62, no.~1, pp.~1--62, 2022.

\bibitem{worldsurvey}
Y.~Guan, H.~Liao, and et~al., ``World models for autonomous driving: An initial survey,'' {\em IEEE Transactions on Intelligent Vehicles}, 2024.

\bibitem{magicdrivedit}
R.~Gao, K.~Chen, and et~al., ``Magicdrivedit: High-resolution long video generation for autonomous driving with adaptive control,'' {\em arXiv preprint arXiv:2411.13807}, 2024.

\bibitem{magicdrive}
R.~Gao, K.~Chen, and et~al., ``Magicdrive: Street view generation with diverse 3d geometry control,'' in {\em ICLR}, 2024.

\bibitem{vista}
S.~Gao, J.~Yang, and et~al., ``Vista: A generalizable driving world model with high fidelity and versatile controllability,'' {\em NIPS}, vol.~37, pp.~91560--91596, 2025.

\bibitem{drive-WM}
Y.~Wang, J.~He, and et~al., ``Driving into the future: Multiview visual forecasting and planning with world model for autonomous driving,'' in {\em CVPR}, pp.~14749--14759, 2024.

\bibitem{xie2025glad}
B.~Xie, Y.~Liu, and et~al., ``Glad: A streaming scene generator for autonomous driving,'' in {\em ICLR}, 2025.

\bibitem{streamingt2v}
R.~Henschel, L.~Khachatryan, and et~al., ``Streamingt2v: Consistent, dynamic, and extendable long video generation from text,'' {\em arXiv preprint arXiv:2403.14773}, 2024.

\bibitem{qian2024nuscenes}
T.~Qian, J.~Chen, and et~al., ``Nuscenes-qa: A multi-modal visual question answering benchmark for autonomous driving scenario,'' in {\em AAAI}, vol.~38, pp.~4542--4550, 2024.

\bibitem{bevcontrol}
K.~Yang, E.~Ma, and et~al., ``Bevcontrol: Accurately controlling street-view elements with multi-perspective consistency via bev sketch layout,'' {\em arXiv preprint arXiv:2308.01661}, 2023.

\bibitem{BEVGen}
A.~Swerdlow, R.~Xu, and et~al., ``Street-view image generation from a bird's-eye view layout,'' {\em IEEE Robotics and Automation Letters}, 2024.

\bibitem{gaia}
A.~Hu, L.~Russell, and et~al., ``Gaia-1: A generative world model for autonomous driving,'' {\em arXiv preprint arXiv:2309.17080}, 2023.

\bibitem{genad}
J.~Yang, S.~Gao, and et~al., ``Generalized predictive model for autonomous driving,'' in {\em CVPR}, pp.~14662--14672, 2024.

\bibitem{wang2023drivedreamer}
X.~Wang, Z.~Zhu, and et~al., ``Drivedreamer: Towards real-world-drive world models for autonomous driving,'' in {\em ECCV}, pp.~55--72, Springer, 2024.

\bibitem{drivingdiffusion}
X.~Li, Y.~Zhang, and et~al., ``Drivingdiffusion: Layout-guided multi-view driving scenarios video generation with latent diffusion model,'' in {\em ECCV}, pp.~469--485, Springer, 2024.

\bibitem{panacea}
Y.~Wen, Y.~Zhao, and et~al., ``Panacea: Panoramic and controllable video generation for autonomous driving,'' in {\em CVPR}, 2024.

\bibitem{DIT}
W.~Peebles and S.~Xie, ``Scalable diffusion models with transformers,'' in {\em CVPR}, pp.~4195--4205, 2023.

\bibitem{dreamforge}
J.~Mei, Y.~Ma, and et~al., ``Dreamforge: Motion-aware autoregressive video generation for multi-view driving scenes,'' in {\em ECCV Workshop}.

\bibitem{latent}
Y.~He, T.~Yang, and et~al., ``Latent video diffusion models for high-fidelity long video generation,'' {\em arXiv preprint arXiv:2211.13221}, 2022.

\bibitem{NUWA-XL}
S.~Yin, C.~Wu, and et~al., ``Nuwa-xl: Diffusion over diffusion for extremely long video generation,'' in {\em ACL}, 2023.

\bibitem{longvideogen}
T.~Brooks, J.~Hellsten, and et~al., ``Generating long videos of dynamic scenes,'' in {\em NeurIPS}, 2022.

\bibitem{fifo}
J.~Kim, J.~Kang, and et~al., ``Fifo-diffusion: Generating infinite videos from text without training,'' in {\em NeurIPS}, 2024.

\bibitem{rombach2022high}
R.~Rombach, A.~Blattmann, and et~al., ``High-resolution image synthesis with latent diffusion models,'' in {\em CVPR}, pp.~10684--10695, 2022.

\bibitem{chengeodiffusion}
K.~Chen, E.~Xie, and et~al., ``Geodiffusion: Text-prompted geometric control for object detection data generation,'' in {\em ICLR}.

\bibitem{hu2022model}
A.~Hu, G.~Corrado, and et~al., ``Model-based imitation learning for urban driving,'' {\em NIPS}, vol.~35, pp.~20703--20716, 2022.

\bibitem{karras2022elucidating}
T.~Karras, M.~Aittala, and et~al., ``Elucidating the design space of diffusion-based generative models,'' {\em NIPS}, vol.~35, pp.~26565--26577, 2022.

\bibitem{FVD}
T.~Unterthiner, S.~van Steenkiste, and et~al., ``Towards accurate generative models of video: A new metric \& challenges,''

\bibitem{FID}
M.~Heusel, H.~Ramsauer, and et~al., ``Gans trained by a two time-scale update rule converge to a local nash equilibrium,'' in {\em NeurIPS}, pp.~6629--6640, 2017.

\bibitem{Videogpt}
W.~Yan, Y.~Zhang, and et~al., ``Videogpt: Video generation using vq-vae and transformers,'' {\em arXiv preprint arXiv:2104.10157}, 2021.

\bibitem{inception-v3-model}
C.~Szegedy, V.~Vanhoucke, and et~al., ``Rethinking the inception architecture for computer vision,'' in {\em CVPR}, pp.~2818--2826, 2016.

\bibitem{blattmann2023stable}
A.~Blattmann, T.~Dockhorn, and et~al., ``Stable video diffusion: Scaling latent video diffusion models to large datasets,'' {\em arXiv preprint arXiv:2311.15127}, 2023.

\end{thebibliography}
